\documentclass[conference]{IEEEtran}
\IEEEoverridecommandlockouts

\usepackage{cite}
\usepackage{amsmath,amssymb,amsfonts}
\usepackage{algorithmic}
\usepackage{upgreek}
\usepackage{graphicx}
\usepackage{textcomp}
\usepackage{xcolor}
\usepackage{algorithm2e}
\SetKwComment{Comment}{/* }{ */} 
\DeclareMathOperator*{\argmin}{arg\,min}
\graphicspath{{.}} 
\usepackage{bm} 
\usepackage{microtype} 
\usepackage{hyperref} 
\hypersetup{
	colorlinks=true,
	linkcolor=blue,
	citecolor=blue,
	urlcolor=blue
}
\usepackage{scalerel}
\usepackage{tikz}
\usetikzlibrary{svg.path}
\definecolor{orcidlogocol}{HTML}{A6CE39} 
\tikzset{
	orcidlogo/.pic={
		\fill[orcidlogocol] (0,0) circle [radius=0.16]; 
		\node[text=white,scale=0.5] at (0,0) {%
			\textsf{\textbf{ID}}}; 
	}
}
\newcommand{\orcidicon}[1]{%
	\href{https://orcid.org/#1}{
		\begin{tikzpicture}[scale=0.5]
			\pic{orcidlogo}; 
		\end{tikzpicture}%
	}%
}

\newcommand{\IEEECopyrightHeader}{
	\scriptsize
	\begin{minipage}[b]{\textwidth}
		This is the accepted manuscript of the paper published in \textit{2025 IEEE 101st Vehicular Technology Conference (VTC2025-Spring)}, \copyright~IEEE, 2025. Received (Best Conference Paper) Award. The final published version is available at \url{https://doi.org/10.1109/VTC2025-SPRING65109.2025.11174939}.
	\end{minipage}
}

\newcommand{\IEEECopyrightFooter}{
	\scriptsize
	\begin{minipage}[t]{\textwidth}
		\textbf{IEEE Copyright Notice:} \copyright~2025 IEEE.  Personal use of this material is permitted.  Permission from IEEE must be obtained for all other uses, in any current or future media, including reprinting/republishing this material for advertising or promotional purposes, creating new collective works, for resale or redistribution to servers or lists, or reuse of any copyrighted component of this work in other works.
	\end{minipage}
}

\makeatletter
\def\ps@IEEEtitlepagestyle{
	\def\@oddhead{\IEEECopyrightHeader}
	\def\@evenhead{\IEEECopyrightHeader}
	\def\@oddfoot{\IEEECopyrightFooter}
	\def\@evenfoot{\IEEECopyrightFooter}
}
\def\ps@headings{
}
\makeatother

\begin{document}
	
	\title{Safety-oriented Dynamic Path Planning for Automated Vehicles\\
		\thanks{This research paper is part of [project MORE -- Munich Mobility Research Campus] and funded by dtec.bw -- Digitalization and Technology Research Center of the Bundeswehr. dtec.bw is funded by the European Union -- NextGenerationEU.}}
	
	\author{
		\IEEEauthorblockN{Mostafa Emam\orcidicon{0000-0003-4942-1183}}
		\IEEEauthorblockA{
			\textit{Department of Aerospace Engineering,}\\
			\textit{University of the Bundeswehr Munich,}\\
			Neubiberg, Munich, Germany\\
			mostafa.emam@unibw.de
		}
		\and
		\IEEEauthorblockN{Matthias Gerdts\orcidicon{0000-0001-8674-5764}}
		\IEEEauthorblockA{
			\textit{Department of Aerospace Engineering,}\\
			\textit{University of the Bundeswehr Munich,}\\
			Neubiberg, Munich, Germany\\
			matthias.gerdts@unibw.de
		}
	}
	
	\maketitle
	
	\begin{abstract}
		Ensuring safety in autonomous vehicles necessitates advanced path planning and obstacle avoidance capabilities, particularly in dynamic environments. This paper introduces a bi-level control framework that efficiently augments road boundaries by incorporating time-dependent grid projections of obstacle movements, thus enabling precise and adaptive path planning. The main control loop utilizes Nonlinear Model Predictive Control (NMPC) for real-time path optimization, wherein homotopy-based constraint relaxation is employed to improve the solvability of the optimal control problem (OCP). Furthermore, an independent backup loop runs concurrently to provide safe fallback trajectories when an optimal trajectory cannot be computed by the main loop within a critical time frame, thus enhancing safety and real-time performance. Our evaluation showcases the benefits of the proposed methods in various driving scenarios, highlighting the real-time applicability and robustness of our approach. Overall, the framework represents a significant step towards safer and more reliable autonomous driving in complex and dynamic environments.
	\end{abstract}
	

	
	\section{Introduction}
	Safety is a paramount concern in the development of Autonomous Vehicles (AVs) and is particularly studied in their three traditional application fields: Aerial (UAV) \cite{c0064}, Ground (AGV) \cite{c0054}, and Surface (ASV) vessels \cite{c0039}. Though a general consensus on the definition of safety is lacking in the literature, there exists some guidelines and legal requirements, from which quantifiable metrics can be derived to determine whether the behavior of an AV is safe or not \cite{c0071,c21}.
	
	In scope of Automated Driving (AD), safety encompasses compliance with traffic laws (e.g., right-of-way), collision avoidance, and promoting passenger comfort and/or cargo safety, especially in dynamic and uncertain environments. This can be formulated as a multi-objective path planning problem, for which different algorithms and solution paradigms exist \cite{c0054}. A critical challenge lies in real-time obstacle avoidance, particularly under constraints imposed by dynamically changing road geometries and the movement of surrounding obstacles \cite{c0060}. To address this challenge, AVs must reliably plan collision-free trajectories that can adapt to rapidly changing conditions and respond to unexpected events, such as erratic behaviors from other traffic participants.
	
	In our previous work \cite{c0065}, we proposed an efficient real-time obstacle avoidance method employing multi-objective Nonlinear Model Predictive Control (NMPC) and the semi-smooth Newton method. This approach yielded optimal driving trajectories and demonstrated computational efficiency and control action smoothness, yet it relied on fixed lane boundaries and did not fully account for dynamic, time-dependent obstacle constraints. To address these limitations, the current paper introduces a framework that integrates curve fitting using natural cubic splines to dynamically augment lane boundaries and account for multiple moving obstacles. By incorporating time-dependent grid projections of obstacle movements, we enable more precise and adaptive path planning. Furthermore, we promote safety by introducing a bi-level control architecture, which enables the generation of safe, fallback trajectories as an independent backup strategy. If an optimal trajectory cannot be found within a time limit, the AV can always follow the fallback trajectory, which guarantees safety and real-time applicability.

	
	\section{Related Work}
	We begin with delineating our objectives for AV control: the ego-vehicle is to track a predetermined path while optimizing for multiple performance criteria, such as minimizing deviations from the reference path, reducing travel time, and enhancing passenger comfort. This shall occur within the bounds of operational and environmental constraints, including the vehicle dynamic limits and collision avoidance. These objectives can be formulated as a traditional multi-objective Model Predictive Contouring Control (MPCC) problem \cite{c0067,c0069}, which is also known in literature as path following in curvilinear coordinates \cite{c0014,c0072} or in the Frenet frame \cite{c0063,c0061}. In addition, collision avoidance can be incorporated using smooth representations, e.g., using Control Barrier Functions (CBF) \cite{c0016}, Gaussians \cite{c0069}, or sigmoid functions \cite{c0065}. This combination ensures safety guarantees while maintaining a relatively low complexity level for the Optimal Control Problem (OCP).
	
	Another aspect is identifying safety and safe vehicle states. \cite{c0076} proposes quantifying the risk related to colliding with different traffic participants (e.g., pedestrians and trucks) by constructing a 2D risk map to represent the collision risk with existing entities in a given driving situation. Afterwards, a bi-level OCP is solved using Sequential Quadratic Programming (SQP) to determine the path with minimum risk and steering effort, which guarantees safety through collision avoidance. While the method is proven effective, it does not prioritize a reference path in the absence of collision threats and requires adaptation to dynamic obstacles. Moreover, no information was given on its real-time applicability. \cite{c0075} follows a similar approach by employing a repulsive potential field to build a 2D map of safe and risky regions in an obstacle-laden environment. This serves as the base grid for a modified A*-based waypoint generation algorithm that finds a smooth, collision-free path on the grid, which is then tracked using a performance-based, fault-tolerant controller. The proposed approach is compared against different adaptations of the A* algorithm and tracking controllers to prove its robustness and efficiency. However, two main drawbacks are the absence of reference path prioritization and the necessary adaptations for handling dynamic obstacles. Furthermore, the developed model-dependent controller makes it difficult to accommodate multi-objective control goals. 
	
	A more relevant A*-based algorithm extension is studied in \cite{c0059}, wherein kinematically feasible paths are constructed using continuous clothoid arcs with linear velocity profiles. Collision avoidance is guaranteed through A*-based path generation, and a multi-objective heuristic cost is utilized for node expansion and selection. The algorithm was further extended in \cite{c0060} to accommodate two additional node expansion strategies: One-Shot and Pure Pursuit, which were implemented and tested on the autonomous vehicle MuCAR-3. All strategies exhibit accurate path following and promising real-time capabilities, with the advantage of returning a sub-optimal solution if a desired time limit has elapsed. Nevertheless, a major disadvantage of the search-based method is that poor planning of a coarse A* trajectory can prevent the subsequent optimization stage from finding any feasible solution. Additionally, it is challenging to directly apply the vehicle dynamic constraints.
	
	\cite{c0071,c0072} introduce the notion of Invariably Safe States and Fail-safe trajectory planning, in which strategies for both braking and steering maneuvers are explored to achieve safe, collision-free routes. Notably, the concept of safe states is extended in \cite{c0071} to an infinite time horizon, elevating its relevance for optimal control strategies like MPC to exceed the immediate control horizon. Despite its real-time efficacy, the strategy's reliance on lane-based maneuvers restricts its utility in unstructured or lane-free environments. Furthermore, the decoupled nature of the bi-level problem in \cite{c0072} implies the existence of abrupt changes of constraints during lane switching, an issue that remains unaddressed and may lead to problem infeasibility.
	
	Finally, an alternative paradigm is explored by \cite{c0074}, in which a minimalist Autonomous Emergency Braking System (AEBS) is implemented and deployed on a micro-controller for real-time responsiveness. The system computes a dynamic safety region from the desired travel path and vehicle speed, then utilizes raw LiDAR data to perform an emergency braking maneuver in the event of an imminent collision. Recognizing that it does not replace the path planning controller, we draw inspiration from it to devise our fallback trajectories, serving as a safeguard to halt the vehicle in case an optimal trajectory cannot be computed within a critical time frame.

	\section{Methodology}
	This manuscript is a continuation of \cite{c0065}. In the interest of brevity, we direct readers to it for an in-depth understanding of the OCP and the determination of in-lane obstacle protrusions, among other topics. In the sequel, we interchangeably use the terms \textit{road edges}, \textit{lane boundaries}, and \textit{tunnel boundaries} to denote the admissible driving area. The primary contributions of this paper are as follows:
	
	\begin{itemize}
		\item Redefining the control problem with a local prediction horizon to better suit the input data streams from vehicle sensors. This data is processed to identify the preliminary admissible driving tunnel.
		\item Accounting for dynamic obstacles by associating them with adequate motion models and projecting their movement unto a time-based grid. This is employed to augment the original tunnel boundaries to ensure smooth and obstacle-free travel paths.
		\item Introducing a continuously differentiable tunnel-width function for road blockade detection and speed control.
		\item Proposing a homotopy-based relaxation approach to improve the OCP solvability.
		\item Computing fail-safe fallback trajectories that the AV can follow in case of the OCP controller timeout.
	\end{itemize}
	
	We now address each of these points in further detail.
	
	\subsection{Path Planning Problem}
	With the assumption of sufficient AV hardware and software components \cite{c0054,c0075}, we expect that each time the controller is invoked, there exist sequenced lists of equidistant data points in Cartesian coordinates denoting the reference (desired) path and the boundaries of the driving area, i.e., the rightmost and leftmost road edges. The points are amenable to transformation into continuously differentiable paths and we use a 2D natural cubic spline to construct $\gamma_r(s) : [0, L] \rightarrow \mathbb{R}^2$ from the reference path data points, which delineates the Frenet frame coordinates. Subsequently, the lane boundaries' data points are projected unto $\gamma_r$ to create the 1D splines $\{\underline{\gamma}(s), \overline{\gamma}(s)\} : [0, L] \rightarrow \mathbb{R}$ for the right and left boundaries, respectively. $\underline{\gamma}, \overline{\gamma}$ denote the boundaries' lateral deviations with respect to arclength $s$, and are used in conjunction with $\gamma_r$ to offer an approximate local projection of any point unto a lane boundary (and vice versa).

	We utilize the linearized point-mass kinematic vehicle model in Frenet frame, which combines simplicity and modularity to facilitate path planning tasks for AVs \cite{c0038}. It describes the curvilinear movement of a point $\rho$ (e.g., the midpoint of the rear axle) on the parameterized curve $\gamma_r$ as
	\begin{equation}\label{eqn:lin_sys_dyn}
		\begin{split}
			&s'(t) = v(t)\\
			&d'(t) = v(t) \chi(t)\\
			&\chi'(t) = v(t) (\kappa(t) - \kappa_r(s(t)))\\
			&\kappa'(t) = u_1(t)\\
			&v'(t) = u_2(t)
		\end{split}
	\end{equation}
	where $s$ is the arclength of the projection of $\rho$ unto $\gamma_r$ and $d$ is its lateral offset. $\chi$ is the heading difference between the AV and $\gamma_r$ at $\rho$, $\kappa$ is the AV's curvature, and $v$ is its velocity. This constitutes the state vector $x=(s,d,\chi,\kappa,v)^T$. The generic inputs $u = (u_1,u_2)^T$ represent the derivative of the curvature and the acceleration, respectively, and can be converted into the controls of different vehicle models with a proper mapping \cite{c0065}. For example, $u_1$ directly sets $\kappa$ in (\ref{eqn:lin_sys_dyn}), but $\kappa$ may be computed from the AV's yaw rate $\varphi'(t)$ with $\kappa(t) = \frac{\varphi'(t)}{v(t)}$. Finally, the curvature of a curve parameterized with respect to arclength is
	\begin{equation*}
		\kappa_r(s) = x'(s) y''(s) - x''(s)y'(s)
	\end{equation*}

	Model (\ref{eqn:lin_sys_dyn}) is subject to the nonlinear constraints
	\begin{equation} \label{eqn:mpc_cnstr}
		\begin{split}
			s(t) &\le L\\
			\underline{d}(s(t)) \le d(t) &\le \overline{d}(s(t))\\
			0 \le v(t) &\le \overline{v}(s(t))\\
			-\overline{a_n} \le \kappa(t)v(t)^2 &\le \overline{a_n}\\
			u \in U
		\end{split}
	\end{equation}
	where $s \le L$ ensures that the AV does not exceed the defined path length and $v \ge 0$ maintains forward progression. Obstacle avoidance is achieved by restricting the AV's lateral deviation within a permissible area $d \in [\underline{d}(s),\overline{d}(s)]$ and its speed below a safe velocity $v \le \overline{v}(s)$. These constraints will be later discussed in detail. Also, vehicle-dependent limits on lateral acceleration $\kappa v^2 \in [-\overline{a_n},\overline{a_n}]$ and system controls $u \in U$ prevent unsafe maneuvers and guarantee feasible and comfortable travel paths. To summarize, the OCP on the time horizon $[0, T]$ reads

	\emph{ Minimize
		\begin{multline*}
			- \alpha_s s(T) + \int_0^{T} \alpha_d {d(t)}^2 + \alpha_{\chi} {\chi(t)}^2 + \alpha_{u_1} {u_1(t)}^2 + \alpha_{u_2} {u_2(t)}^2  dt
		\end{multline*}
		\indent s.t. (\ref{eqn:lin_sys_dyn}), (\ref{eqn:mpc_cnstr}), and the initial values $x(0) = x_0$.
	}\\[2pt]
	where the objective function entails maximizing the traveled distance $s\left(T\right)$ while minimizing the tracking errors $d, \chi$ and the system controls $u$.
	
	\subsection{Incorporating Obstacle Avoidance}
	Proceeding with the previous assumption, we expect that the system states (e.g., position and velocity) of all present obstacles are available to the MPC. From a bird's-eye view, we presume that the 2D footprint of any obstacle $o$ is encapsulated by a set of unordered points $O_{xy}$. $o$ can be associated with an appropriate motion model $M(o) :=  \{x'(t) = f(x(t), u(t))\}$, such as Constant Velocity (CV) for pedestrians and Constant Curvature and Acceleration (CCA) for vehicles \cite{c0077}. By applying the Runge-Kutta (RK4) iterative method to $M(o)$, we can predict the time-dependent obstacle positions over the interval $[0, T]$ with a sampling time of $\delta$ and $N$ discrete steps, where $T=\delta N$. Using the obstacle data $O_{xy}, M(o)$, the prediction variables $\delta, N$, and the path splines $\gamma_r, \underline{\gamma}, \overline{\gamma}$, we can determine the in-lane time-dependent obstacle protrusions $P_{sd}(k) := \{(s_i,d_i) \,|\, i \in \mathbb{N}_0, k = 0, 1, \ldots, N\}$ clipped at the nearest boundary $\eta$. The process is depicted in Algorithm (\ref{alg:proj_obstacle}).

	\begin{algorithm}
		\caption{Determine the the nearest boundary $\eta$ and the in-lane time-dependent obstacle protrusions $P_{sd}(k)$}\label{alg:proj_obstacle}
		\KwIn{$O_{xy}, M(o), \delta, N, \gamma_r, \underline{\gamma}, \overline{\gamma}$}
		\KwOut{$\eta, P_{sd}(k)$}
		$\tilde{O}_{xy} \gets refineXY(O_{xy})$\;
		$O_{sd}(0) \gets transformXYtoSD(\gamma_r, \tilde{O}_{xy})$\;
		\For{$k=0$ \KwTo $N - 1$}
		{
			$O_{sd}(k+1) \gets predictStates(O_{sd}(k), M(o), \delta)$\;
		}
		$\overline{O}_{sd}(0) \gets localProjection(\gamma_r, \overline{\gamma}, O_{sd}(0))$\;
		$\underline{O}_{sd}(0) \gets localProjection(\gamma_r, \underline{\gamma}, O_{sd}(0))$\;
		$d_l \gets max{\{d_i \,|\, (s_i,d_i)\in \overline{O}_{sd}(0)\}}$\;
		$d_r \gets min{\{d_i \,|\, (s_i,d_i)\in \underline{O}_{sd}(0)\}}$\;
		\eIf{$((d_r <= 0) \land (d_l <= 0)) \lor (d_r \le -d_l)$}
		{
			$\eta \gets -1$\Comment*[r]{Align right}
			\For{$k=0$ \KwTo $N$}
			{
				$\underline{O}_{sd}(k) \gets localProjection(\gamma_r, \underline{\gamma}, O_{sd}(k))$\;
				$\underline{P}_{sd}(k) \gets sutherlandHodgman(\underline{\gamma}, \underline{O}_{sd}(k))$\;
				$P_{sd}(k) \gets localProjection(\underline{\gamma}, \gamma_r, \underline{P}_{sd}(k))$\;
			}
		}{
			$\eta \gets 1$\Comment*[r]{Align left}
			\For{$k=0$ \KwTo $N$}
			{
				$\overline{O}_{sd}(k) \gets localProjection(\gamma_r, \overline{\gamma}, O_{sd}(k))$\;
				$\overline{P}_{sd}(k) \gets sutherlandHodgman(\overline{\gamma}, \overline{O}_{sd}(k))$\;
				$P_{sd}(k) \gets localProjection(\overline{\gamma}, \gamma_r, \overline{P}_{sd}(k))$\;
			}
		}
	\end{algorithm}

	The workflow can be explained as follows:
	\begin{itemize}
		\item \textit{Refine Input}: Make a convex hull from $O_{xy}$ with Graham's scan \cite{c0056} to arrange the data points and remove redundant ones. Then, augment it with additional points based on a minimum inter-point distance to yield $\tilde{O}_{xy}$. This preserves the obstacle shape after coordinate transformation.
		\item \textit{Transform Input}: Project the ordered, augmented points $\tilde{O}_{xy}$ unto $\gamma_r$ and store them as the initial obstacle position in Frenet frame $O_{sd}(0)$.
		\item \textit{Predict States}: Using $O_{sd}(0)$ and $M(o)$, predict the positions $O_{sd}(k+1)$ at the discrete steps $k = 0, 1, \ldots, N-1$. For efficiency, we associate $M(o)$ with a singular point $O_{sd,m}(k) := \{(s_m, d_m) \in O_{sd} \,|\, \forall k\}$ at rotation $\theta_m$. Then, we find the polar relationships $r_i, \theta_i$ for each point in $O_{sd}$ relative to it, wherein $r_i := \sqrt{(s_i - s_m)^2 + (d_i - d_m)^2}$ and $\theta_i := \tan^{-1}(\frac{d_i - d_m}{s_i - s_m})$. Accordingly, we use $M(o)$ to predict $O_{sd,m}(k+1), \theta_m(k+1)$ and employ this to update the values of each other point in $O_{sd}$ with the relationships $s_i (k+1) = s_m (k+1) + r_i \cos \left( \theta_i + \theta_m(k+1) \right)$ and $d_i (k+1) = d_m (k+1) + r_i \sin \left( \theta_i + \theta_m (k+1) \right)$.
		\item \textit{Get Nearest Edge}: With $\underline{\gamma}, \overline{\gamma}$, get the local projection of $O_{sd}(0)$ unto both road edges to yield $\underline{O}_{sd}(0), \overline{O}_{sd}(0)$ for the right and left boundaries, respectively. Based on the minimum lateral distance to the nearest edge, set the value $\eta \in \{-1: right, 1 :left\}$, which remains unchanged throughout the current MPC iteration for solution stability.
		\item \textit{Get Protrusions}: Over $N$ time steps, locally project $O_{sd}(k)$ unto the nearest edge, then use a single iteration of Sutherland-Hodgman (SH) clipping \cite{c0055} to identify the in-lane protrusions. For instance: clip $\underline{O}_{sd}(0)$ at $\underline{\gamma}$ to get $\underline{P}_{sd}(0)$. Finally, apply a reverse local projection to $\gamma_r$ to find the in-lane protrusions in Frenet frame $P_{sd}(k)$.
	\end{itemize}
	Despite the absence of dedicated benchmarking efforts, we propose that Algorithm \ref{alg:proj_obstacle} is sufficiently efficient. For instance, processing an obstacle of $|O_{xy}| = 8$ with $\delta = 0.15[s], N = 30$ and a varying $|P_{sd}(k)|$ at each time step required, on average, $450 [\upmu s]$. Hence, empirical results suggest the algorithm's practical efficiency without the need for extensive benchmarking.
	
	\subsection{Lane Boundary Augmentation}
	
	We now combine the base splines $\underline{\gamma}, \overline{\gamma}$ with $P_{sd}(k)$ to get the augmented splines $\{\tilde{\underline{\gamma}}(s,k), \tilde{\overline{\gamma}}(s,k)\} : [0, L] \times [0, N] \rightarrow \mathbb{R}$, which denote the boundaries' lateral deviation with respect to arclength $s$ and time step $k$. It starts with extracting the complementary contours $C_{sd}(k)$ of $P_{sd}(k)$, e.g., right contours of left-aligned obstacles, using geometrical exploitation. Afterwards, we compute the extended contours $\tilde{C}_{sd}(k)$ using the obstacle-specific safety margins and the AV dimensions $\{w, l_f, l_b\}$, which are the AV width and lengths with respect to $\rho$. Herein, we use the complementary splines, e.g., right boundary for left-aligned obstacles, to ensure the contour extension yields feasible driving areas. Moreover, the extended contours of proximate obstacles are merged using Graham's scan to prevent excessive maneuvering. Next, we intersect $\tilde{C}_{sd}(k)$ with their corresponding boundary $\underline{\gamma} \text{ or } \overline{\gamma}$ to yield the augmented splines $\tilde{\underline{\gamma}}(s,k), \tilde{\overline{\gamma}}(s,k)$. The total number of boundary splines thus becomes $2(N+1)$ for the augmented and base splines. We highlight here that this procedure manipulates the data points directly instead of the splines for increased computational efficiency. Again, no dedicated benchmarking was conducted; however, the empirical results for processing multiple obstacles under various conditions (including merging operations) took $1850 [\upmu s]$ on average, with a maximum of $2500 [\upmu s]$.

	Instead of abruptly using the augmented splines to define the admissible driving area, we follow a homotopy-based strategy to enhance the OCP tractability and prevent solver failures due to infeasibility \cite{c0078}. We start with a lower complexity problem using only the base splines $\underline{\gamma}(s),\overline{\gamma}(s)$ for the lateral constraints. Then, the problem solution is used as an initial guess for a subsequent step, where the augmented splines $\tilde{\underline{\gamma}}(s,k), \tilde{\overline{\gamma}}(s,k)$ are progressively introduced in relation to $\underline{\gamma}(s),\overline{\gamma}(s)$ using a homotopy index $\zeta^i \in [0, 1]$ as shown in (\ref{eqn:cnstr_hmtp}). This gradually shrinks the permissible driving tunnel and makes the problem stricter until we reach the desired problem at $\tilde{\underline{\gamma}}(s,k), \tilde{\overline{\gamma}}(s,k)$. The number of homotopy iterations $Z$ is a compromise between a faster process and a higher convergence rate per iteration.
	\begin{equation} \label{eqn:cnstr_hmtp}
		\begin{split}
			\underline{d}(s,k,\zeta^i) = (1- \zeta^i) \underline{\gamma}(s) + \zeta^i \tilde{\underline{\gamma}}(s, k)\\
			\overline{d}(s,k,\zeta^i) = (1- \zeta^i) \overline{\gamma}(s) + \zeta^i \tilde{\overline{\gamma}}(s, k)
		\end{split}
	\end{equation}

	In other words, we start with the base splines at $\zeta^1 = 0$. Then, as $i = 1,2,\ldots,Z$, we iteratively increase $\zeta^i$ until $\zeta^Z = 1$, where the constraints are defined solely by $\tilde{\underline{\gamma}}(s,k), \tilde{\overline{\gamma}}(s,k)$. Each iteration $i$ begins with the solution from the preceding step $i - 1$ as its initial guess to leverage the solution convergence.
	
	\subsection{Road Blockade Detection}
	
	A blockade is characterized by the total closure of the driving tunnel as denoted by the overlapping of road boundaries, i.e., when $\overline{d} \le \underline{d}$. Accordingly, we define the tunnel width function $\omega(s,k) := \overline{d}(s,k,\zeta^Z)- \underline{d}(s,k,\zeta^Z)$, in which $\omega(s,k) \le \underline{\omega}$ indicates a road blockade within a specified tolerance $\underline{\omega} \approx 0$. Furthermore, we introduce	
	\begin{equation}
		(s_h, k_h) = \argmin_{s,k} \{(s, k) \mid \omega(s,k) \le \underline{\omega} \, , \,\forall k \}
	\end{equation}
	where $k_h$ is the earliest instance at which $\omega \le \underline{\omega}$ and $s_h$ is the overall closest instance for the same condition. To determine these values, we iterate over all discrete time steps $k$ and apply Newton's method on $\omega$ to find $s_{h,k}$ that satisfies $\omega(s,k) \le \underline{\omega}$. If the condition is not met, a stop is added at the end of the road $s_{h,k} = L$. Subsequently, we take the overall minimum values of $s_{h,k},k_h$ to halt the AV at the earliest possible moment.
	
	For enhanced safety, we employ an anticipatory stop $\tilde{s}_h < s_h$ on the reference path $\gamma_r$ prior to the blockade. This is achieved by backtracking along the protruding boundary at the blockade until $\gamma_r$ is intersected. For example, if the blockade exists due to a right-aligned obstacle, we backtrack from $s_h$ along the right boundary until $\underline{d}(\tilde{s}_h,k,\zeta^Z) \approx \gamma_r(\tilde{s}_h)$ is reached. This guarantees that the AV halts before encountering any potential hazards. Next, we use the admissible controls $U$ to compute the braking distance $s_{decl}$ from $\tilde{s}_h$. The comparison $\tilde{s}_h \ge s_{decl}$ specifies that a comfortable deceleration profile is possible from the road speed limit $\overline{v}_r$. Otherwise, $\tilde{s}_h$ is too close and we compute $\overline{v}_{decl} = f(\tilde{s}_h, U)< \overline{v}_r$ to restrict the max speed at the beginning of the road. 
	
	Finally, we employ the base spline $\overline{v}(s) : [0,L] \rightarrow \overline{v}_r$ and the augmented spline $\tilde{\overline{v}}(s) : [0,L] \rightarrow \mathbb{R}$ with homotopy relaxation
	\begin{equation}
		\begin{split}
			&\tilde{\overline{v}}(s) :=
			\begin{cases}
				\overline{v}_0, & s \in [0, s_{decl})\\
				\frac{\tilde{s}_{h}-s}{\tilde{s}_{h}-s_{decl}}\overline{v}_0, & s \in [s_{decl}, \tilde{s}_h]\\
				0, & s \in (\tilde{s}_h, L]
			\end{cases}\\
			&\overline{v}(s,\zeta^i) = (1 - \zeta^i) \overline{v}(s) + \zeta^i \tilde{\overline{v}}(s)
		\end{split}
	\end{equation}
	where $\overline{v}_{0} := \min(\overline{v}_r, \overline{v}_{decl})$. In conclusion, we revise (\ref{eqn:mpc_cnstr}) to yield the discrete, homotopy-based OCP constraints
	\begin{equation} \label{eqn:mpc_cnstr_hmtp}
		\begin{split}
			s(k) &\le \tilde{s}_h\\
			\underline{d}(s,k,\zeta^i) \le d(k) &\le \overline{d}(s,k,\zeta^i)\\
			0 \le v(k) &\le \overline{v}(s,\zeta^i)\\
			-\overline{a_n} \le \kappa(k)v(k)^2 &\le \overline{a_n}\\
			u \in U
		\end{split}
	\end{equation}
	%
	
	\subsection{Fallback Trajectories}
	
	If the MPC solution is not available within a designated time frame, it is imperative to provide the AV with an alternative trajectory to prevent collisions. Herein, we discuss three possible fallback methods, which guarantee real-time safety.
	
	Let $z_k = (x(k),u(k))^T$ be the system states and controls at step $k$, in which $z_{n,N} = ({z}^T_n,\ldots,{z}^T_{n+N})^T$ denotes the discrete MPC solution at time $n$ over $N$ steps. Let $g(z_{n,N})$ be the constraints defined in (\ref{eqn:mpc_cnstr_hmtp}), where ${nr}_g$ represents the number of constraints and ${nr}_z$ that of system equations. We introduce the function $\texttt{isSafe}(z_{n,N},g) : \mathbb{R}^{{nr}_z} \times \mathbb{R}^{{nr}_g} \rightarrow \{0, 1\}$, which ensures the constraints violation to be within specified tolerances $\epsilon \in \mathbb{R}^{{nr}_g} $. Moreover, it evaluates the path continuity with $s_{n+k} \le s_{n+k+1}, \, k = n,\ldots,n+N-1$ and subsequently ascertains the safety of $z_{n,N}$.
	
	Following homotopy relaxation, we offer the first fallback strategy by using a large number of iterations $Z$, so that the MPC has ample opportunity to find a solution. In case of failure, we validate the last obtained $z_{n,N}$ at $\zeta^i < 1$ using $\texttt{isSafe}$ and adopt this solution if it is deemed feasible.
	
	Second, the MPC paradigm offers a solution over its entire prediction horizon $[0,T]$, yet a single-step approach is typically used by applying only the first index of the control inputs \cite{c0013}. By employing $\texttt{isSafe}$ combined with sensitivity updates, we can compute an approximate solution $\hat{z}_{n,N-N_0}$ for the current time step by extracting a feasible, shifted segment $z_{n-1,N - N_0}$ from the solution of the preceding MPC iteration. The segment is computed using the previous controls $u_{n-1,N}$ and sensitivity matrix $S_{n-1,N}$, and is only defined on the interval $[0, N - N_0]$ due to the segment shift $N_0$. Note that the MPC is designed and solved in Frenet frame, which is redefined at each time step. Accordingly, at the end of each MPC iteration, we transform and store the solution from $n$ in Cartesian coordinates and re-transform it to the new Frenet frame at $n+1$. In addition, we introduce the function $\texttt{isValid}(z_n, z_{n-1,N})$ to determine whether the discrepancies between current system states and previous solution are sufficiently small and can be handled solely by sensitivity updates.
	
	Third, it has been established that a safe state can be generally reached when the AV stops at $\gamma_r$ inside the permissible driving area \cite{c0072}. We can incorporate this into our MPC approach with minimal changes by setting $\zeta = 1, \, \alpha_s = 0$ and introducing $\alpha_v \in \mathbb{R}^+$ to the objective function, where minimizing $\alpha_v v^2$ encourages halting the AV. We denote henceforth this modified problem as ${OCP}_{stop}$. Since this is regarded as an independent task, it can be executed on a separate controller, which operates concurrently with the MPC, to leverage real-time applicability. Likewise, the integration of homotopy constraints can simplify the problem and guarantee the existence of a solution. To conclude, we can follow one of the three fallback strategies on MPC timeout. They return
	
	\begin{itemize}
		\item The latest safe homotopy trajectory from this iteration.
		\item A valid and safe excerpt from the last iteration solution.
		\item A safe trajectory from ${OCP}_{safe}$ to halt the AV.
	\end{itemize}
	
	\section{Evaluation}
	Two practical challenges arise on evaluating typical MPC approaches: operating at dynamic limits and parameter tuning. The first one relates to the MPC acceptance criteria, wherein any solution $z_{n,N}$ is accepted if it respects the prescribed hard constraints $g(z_{n,N})$. However, modeling discrepancies, dynamic scenarios, sensor noise, and, in our case, system transformations between Cartesian and Frenet frames, cause inconsistencies that may yield infeasible system states, which are inherently intractable with MPC. This can be circumvented with penalty functions \cite{c0016} or soft constraints \cite{c0036}, which improve the OCP solvability, but simultaneously increase its complexity and potentially undermine system safety.
	 
	 Instead, we propose adding safety margins $\varepsilon \in \mathbb{R}^{{nr}_g}$ on the hard constraints (\ref{eqn:mpc_cnstr_hmtp}) to get ${OCP}_{safe}$, which has a narrower permissible solution space. These margins can be removed during safety verification stage with $\texttt{isSafe}(z_{n,N},g)$, thus enabling the acceptance of a broader spectrum of solutions. This dual-stage approach essentially forces the MPC to solve a more conservative problem than the actual safety region necessitates and permits the acceptance of the best solution guess, regardless of its infeasibility under the hard constraints of ${OCP}_{safe}$. Also, we propose a basic recovery strategy for the best solution guess, wherein we constrain the system controls within the permissible range with $\hat{u}_n = \min(\max(\underline{u}, u_n), \overline{u})$, then employ $\hat{u}_n$ to update the states over the prediction horizon with $\hat{x}_{n+k+1} = f(x_{n+k}, \hat{u}_n),\, k = n, \ldots, n+N-1$. This, in addition to $\texttt{isSafe}$, ensures the solution feasibility. Moreover, we solve ${OCP}_{stop}$ directly on (\ref{eqn:mpc_cnstr_hmtp}) with $\varepsilon = \{0\}^{{nr}_g}$.
	 
	 Second, since parameter tuning is system specific, we defer this issue to the hardware deployment phase. Nevertheless, it has been established that homotopy relaxation can accommodate different systems without the need for extensive parameter tuning \cite{c0078}, which guarantees the robustness of our approach.
 
 	\subsection{Numerical Testing and Simulation}
 	
 	The controller for solving ${OCP}_{safe}$ and fallback strategies were entirely developed in C++. Herein, we offer five MPC adaptations for a thorough comparison: traditional (single-step) MPC, MPC with basic recovery, homotopy-based MPC, MPC with sensitivity updates, and MPC with ${OCP}_{stop}$ as a fallback strategy. In the sequel, they are denoted as $MPC$, ${MPC}_{RE}$, ${MPC}_{HB}$, ${MPC}_{SU}$, and ${MPC}_{SS}$, respectively.
 	
 	The numerical simulation results were recorded on an Ubuntu 20.04 system with the processor i7-1355U and 32GB of RAM. We employed the trapezoidal rule for OCP discretization and used a prediction horizon of $T = 3.5[s]$ with $N = 25$ control points ($\delta = 0.14[s]$). For ${MPC}_{HB}$ we used $Z=20$ homotopy iterations. The objective function weights were first set using the scaled contribution of system states and controls, then marginally adapted via trial and error. Since operation timeout can be directly implemented on a hardware controller with system interrupts, it is omitted in the simulation environment. Nonetheless, we limit the maximum MPC solution iterations to $30$ to mimic time restrictions. The code and simulation videos are available at: \url{https://doi.org/10.5281/zenodo.14652538}
	\subsection{Static Obstacle Avoidance}
	In the first test scenario, two static obstacle are situated on either sides of $\gamma_r$ in challenging locations to compel the AV to execute aggressive maneuvers for successful navigation. Since all controllers operate on local paths at different time frames, we construct aggregated figures in the sequel for clarity and direct the reader to the repository for the simulation videos.

	\begin{figure}[ht]
		\centering
		\includegraphics[width=0.45\textwidth]{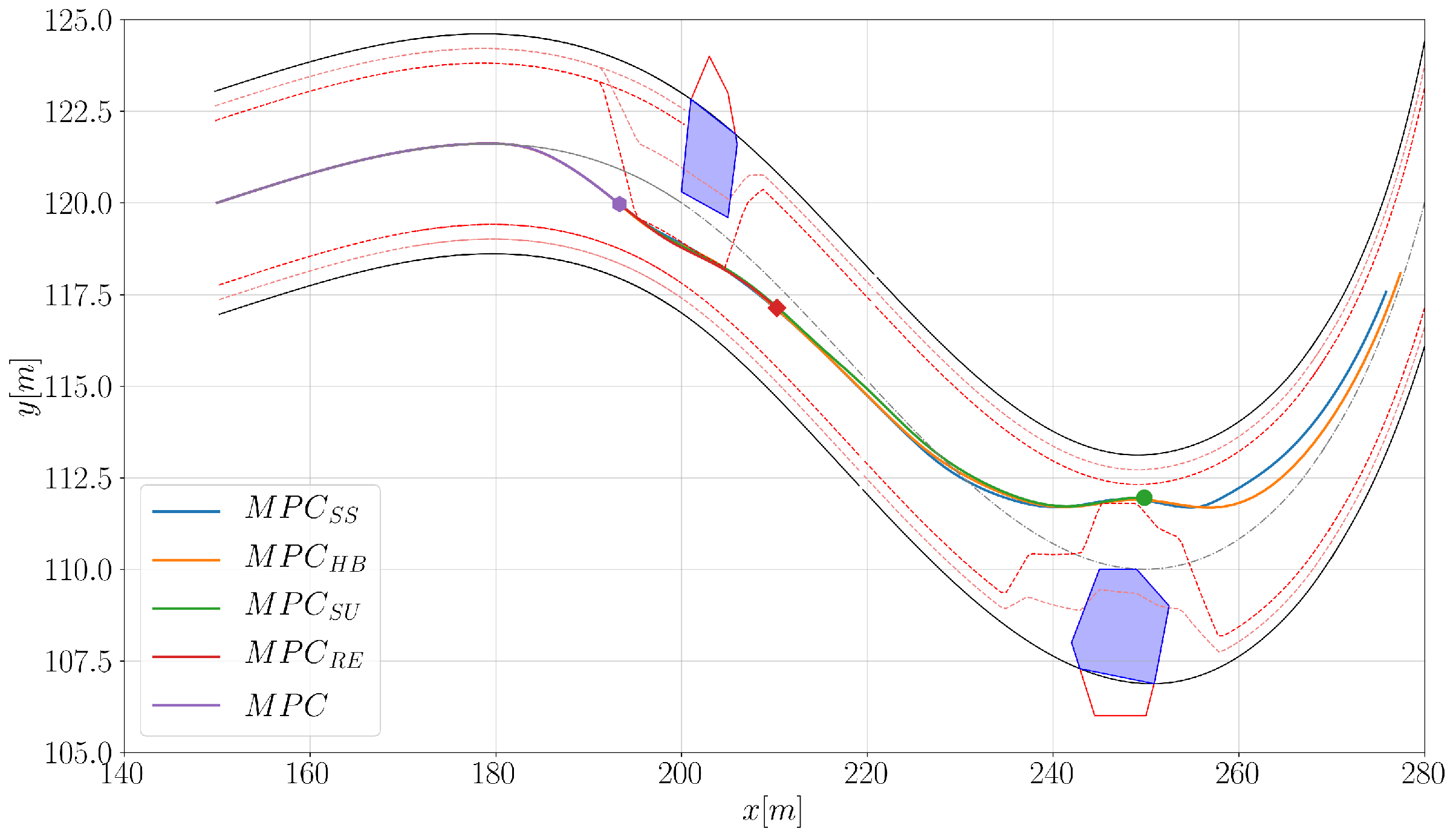}
		\caption{Aggregated constraints and paths for the first test scenario}
		\label{fig:tst_1_path}
	\end{figure}
	Figure (\ref{fig:tst_1_path}) illustrates the different controller paths to traverse the test environment. The road edges are plotted with solid black lines, and the augmented lane boundaries are depicted in red at $\zeta=1.0$ and maroon at $\zeta = 0.5$. The obstacles are represented by red squares, with the identified protrusions being the areas shaded in blue. Only ${MPC}_{\text{HB}}$ and ${MPC}_{\text{SS}}$ managed to successfully complete the maneuver, and the locations, at which the remaining controllers failed, are indicated with markers. This will now be explained in detail.

	$MPC$ struggled with managing dynamic lane boundaries as hard constraints, resulting in an immediate failure at the first instance of operating at the dynamic limit. The problem became infeasible and the controller could not compute a safe trajectory within the permissible number of solving iterations. ${MPC}_{RE}$ fared better, yet the basic recovery strategy proved insufficient to handle the sharp turn required to navigate around the second obstacle. The improved ${MPC}_{SU}$ traversed both turning maneuvers, however, it encountered an impasse at the bottleneck, again owing to the hard constraints.

	\begin{figure}[ht]
		\centering
		\includegraphics[width=0.45\textwidth]{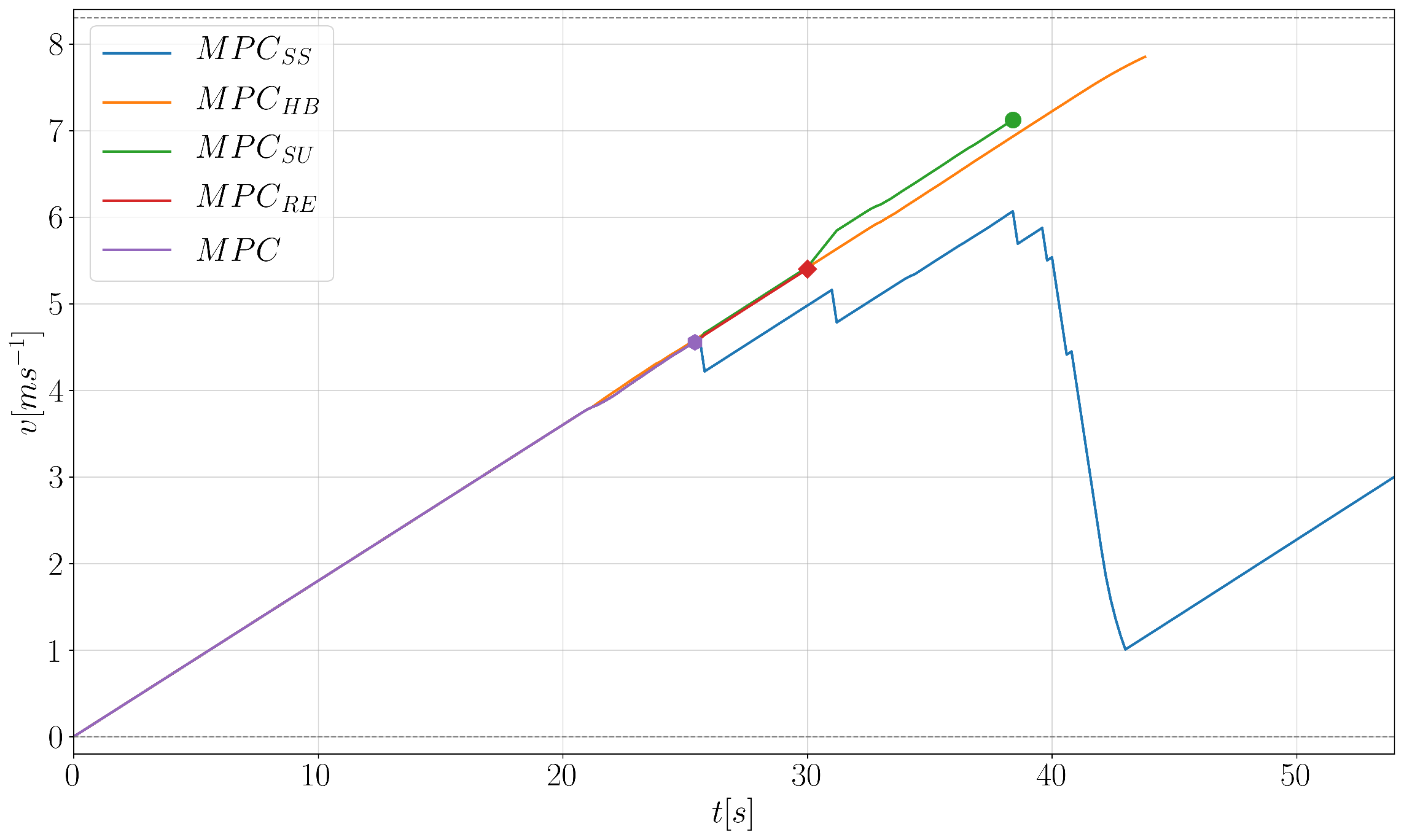}
		\caption{Velocity profiles for the first test scenario}
		\label{fig:tst_1_velocity}
	\end{figure}

	${MPC}_{SS}$ successfully navigated the scenario by frequently engaging the fallback strategy ${OCP}_{stop}$ as shown in Figure (\ref{fig:tst_1_velocity}). The recurring AV deceleration simplified the problem, allowing ${OCP}_{safe}$ to subsequently recover and operate the AV once it had cleared the challenging sections. ${MPC}_{HB}$ displays a clear superiority to the other controllers by not only successfully completing the scenario, but also maintaining a smooth driving trajectory with minimal deceleration.
	
	\subsection{Dynamic Obstacle Overtaking}
\begin{figure}[ht]
		\centering
		\includegraphics[width=0.45\textwidth]{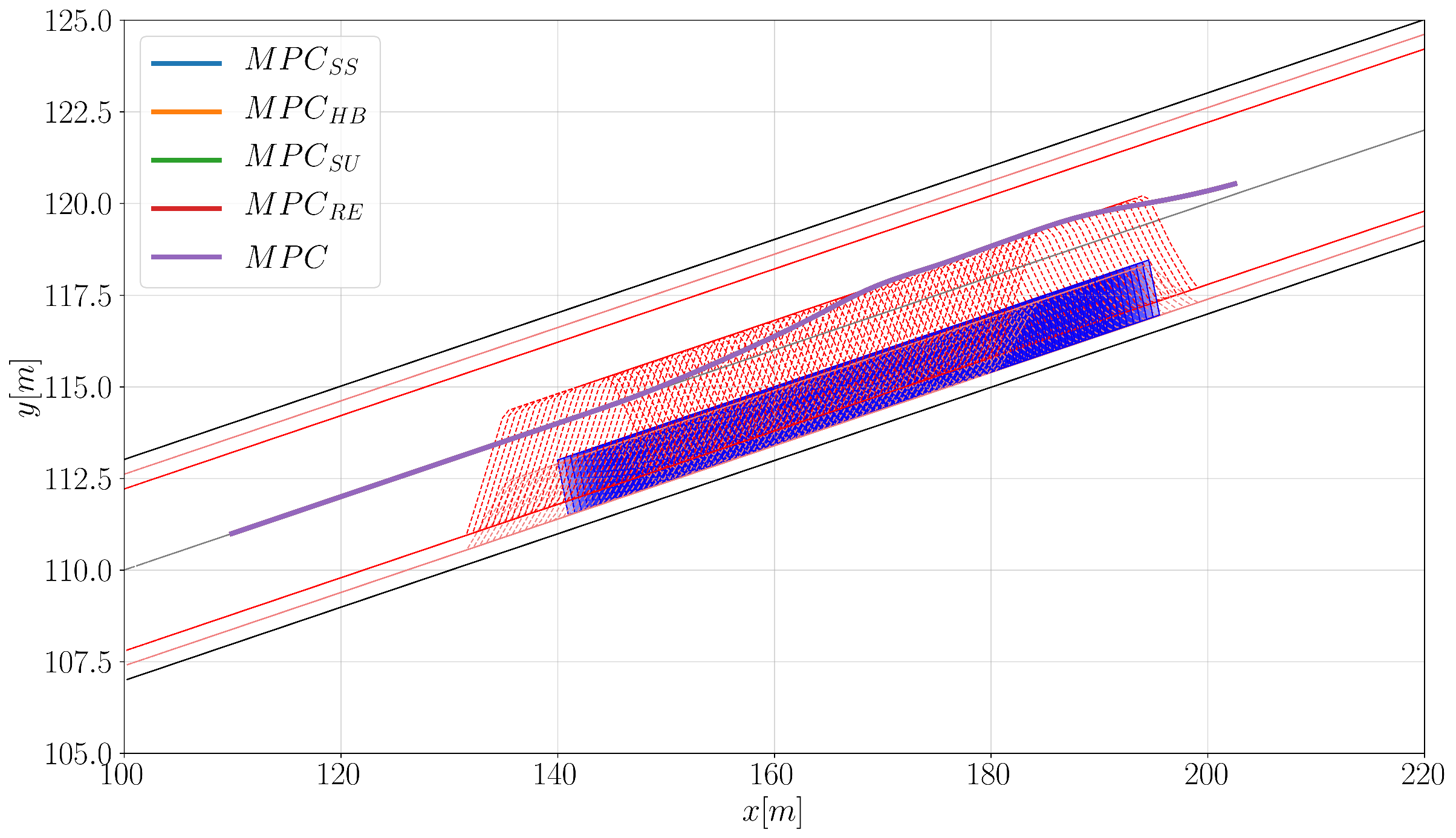}
		\caption{Aggregated constraints and paths for the second test scenario}
		\label{fig:tst_2_path}
	\end{figure}

\begin{figure}[ht]
		\centering
		\includegraphics[width=0.45\textwidth]{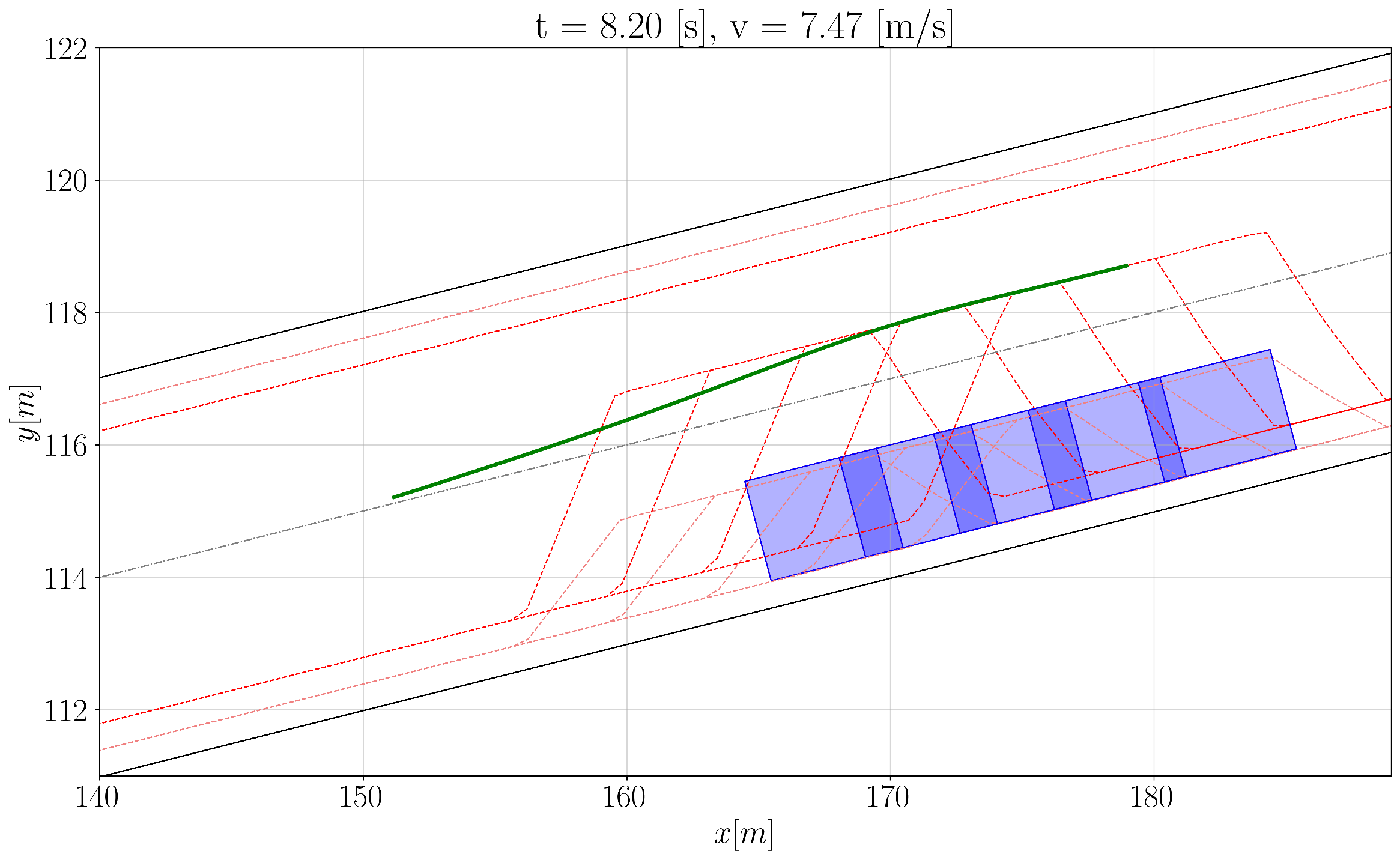}
		\caption{Intermediate ${MPC}_{HB}$ solution at $t=8.2[s]$ and predicted constraints}
		\label{fig:tst_2_t_41}
	\end{figure}

\begin{figure}[ht]
		\centering
		\includegraphics[width=0.45\textwidth]{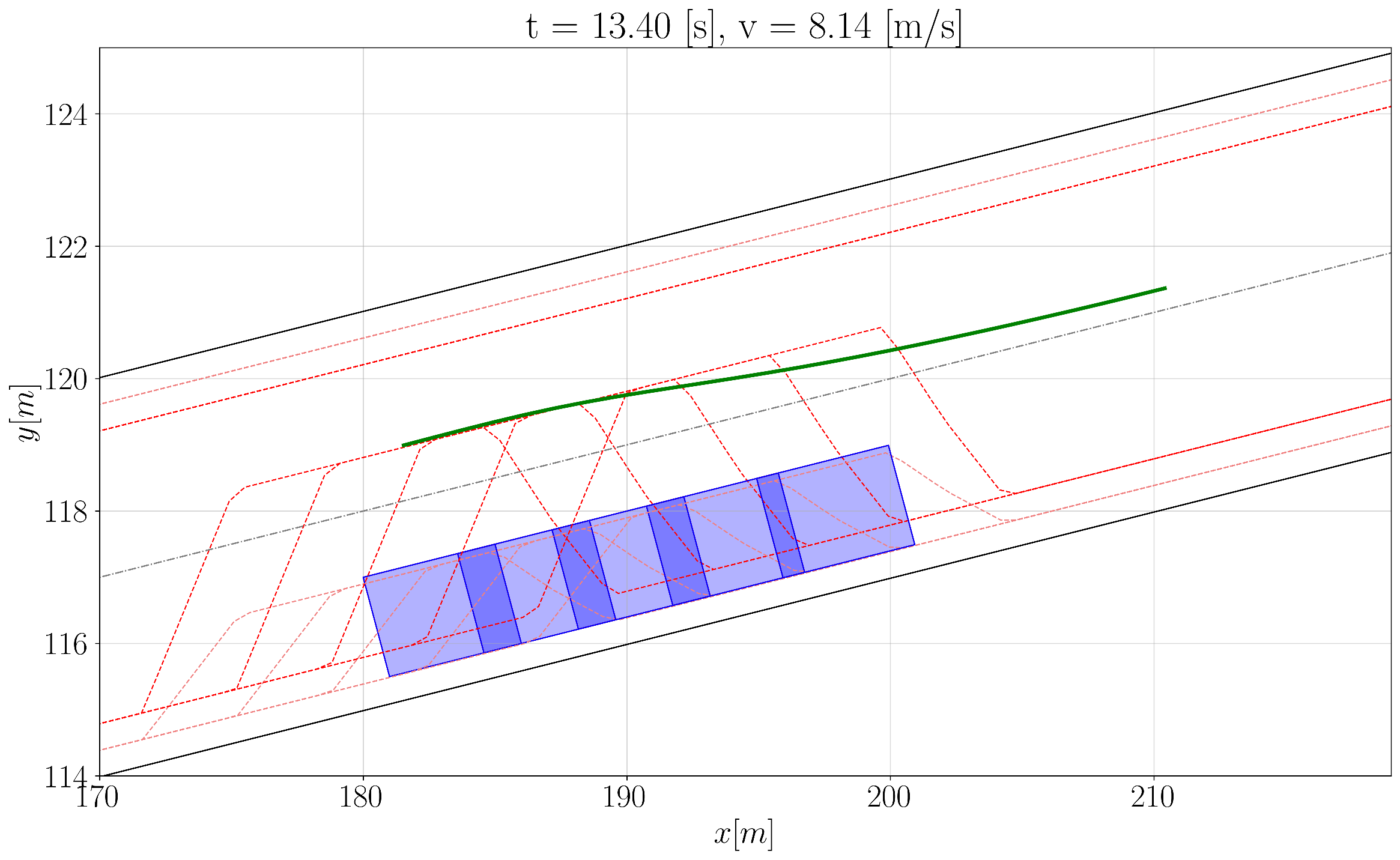}
		\caption{Intermediate ${MPC}_{HB}$ solution at $t=13.4[s]$ and predicted constraints}
		\label{fig:tst_2_t_66}
	\end{figure}

	Next, we evaluate our approach with an overtaking scenario. The test starts with the AV at $v=6.0[ms^{-1}]$ and it aims to overtake a moving obstacle with $v = 4.0[ms^-1]$. Due to the ample maneuvering space, all controllers successfully completed the task with almost indistinguishable travel paths as depicted in Figure (\ref{fig:tst_2_path}). This is an expected outcome due to their shared algorithmic foundation (MPC) with identical objective function weights. Also, the intermediate solutions generated by ${MPC}_{HB}$ are presented in Figures (\ref{fig:tst_2_t_41},\ref{fig:tst_2_t_66}). They illustrate the successful integration of the obstacle's dynamic behavior in the solution computation, while guaranteeing adherence to the problem constraints. Note that both testing scenarios concluded with a slight lateral deviation from $\gamma_r$, which may be readily corrected by parameter tuning of the objective weights.
	\begin{table}[htb]
		\centering
		\caption{Average controller solution time for both test scenarios}
		\label{tbl:tst_exe_mean}
		\begin{tabular}{*5c}
			\hline
			& \multicolumn{2}{c}{Scenario 1} & \multicolumn{2}{c}{Scenario 2}\\
			& $Done?$ & $t_{mean}[ms]$ & $Done?$ & $t_{mean}[ms]$ \\
			\hline
			\hline
			${MPC}$ & $\times$ & $14.311$ & $\checkmark$ & $27.653$\\
			${MPC}_{RE}$ & $\times$ & $19.781$ & $\checkmark$ & $24.472$\\
			${MPC}_{HB}$ & $\checkmark$ & $24.359$ & $\checkmark$ & $35.771$\\
			${MPC}_{SU}$ & $\times$ & $40.917$ & $\checkmark$ & $25.025$\\
			${MPC}_{SS}$ & $\checkmark$ & $44.671$ & $\checkmark$ & $23.873$\\
			${OCP}_{stop}$ & $-$ & $20.096$ & $-$ & $23.141$\\
			\hline
		\end{tabular}
	\end{table}

	Finally, Table (\ref{tbl:tst_exe_mean}) summarizes the test results and lists each controller average computational time. Contrary to \cite{c0078}, we have demonstrated that a modified problem formulation enables using homotopy-based approaches in real-time applications.
	%
	\section{Conclusion}
	In this paper, we present a comprehensive control framework for AVs navigating dynamic environments, with a focus on safety and real-time performance. The key innovation lies in augmenting NMPC with homotopy-based constraints to enhance the problem's tractability and offering an independent backup strategy for fallback trajectory planning. Numerical simulations prove the effectiveness of the proposed methods in challenging and dynamic driving scenarios, and future research will focus on validation on a real-world vehicle.
	
	\section{Acknowledgments}
	The code implemented in this research was facilitated by incorporating the spline library developed by \textit{Andreas Huber} and the OCP solver developed by \textit{Matthias Gerdts}. Both of these tools are in-house efforts at the Department of Aerospace Engineering at the University of the Bundeswehr Munich.

\bibliographystyle{IEEEtran}
\bibliography{IEEEabrv,references}

\begin{thebibliography}{10}
\providecommand{\url}[1]{#1}
\csname url@rmstyle\endcsname
\providecommand{\newblock}{\relax}
\providecommand{\bibinfo}[2]{#2}
\providecommand\BIBentrySTDinterwordspacing{\spaceskip=0pt\relax}
\providecommand\BIBentryALTinterwordstretchfactor{4}
\providecommand\BIBentryALTinterwordspacing{\spaceskip=\fontdimen2\font plus
\BIBentryALTinterwordstretchfactor\fontdimen3\font minus
  \fontdimen4\font\relax}
\providecommand\BIBforeignlanguage[2]{{%
\expandafter\ifx\csname l@#1\endcsname\relax
\typeout{** WARNING: IEEEtran.bst: No hyphenation pattern has been}%
\typeout{** loaded for the language `#1'. Using the pattern for}%
\typeout{** the default language instead.}%
\else
\language=\csname l@#1\endcsname
\fi
#2}}

\bibitem{c0064}
E.~Frazzoli, M.~A. Dahleh, and E.~Feron, ``Real-time motion planning for agile
  autonomous vehicles,'' \emph{Journal of Guidance, Control, and Dynamics},
  vol.~25, no.~1, pp. 116--129, Jan. 2002.

\bibitem{c0054}
H.~Qin, S.~Shao, T.~Wang, X.~Yu, Y.~Jiang, and Z.~Cao, ``Review of autonomous
  path planning algorithms for mobile robots,'' \emph{Drones}, vol.~7, no.~3,
  p. 211, Mar. 2023.

\bibitem{c0039}
A.~Vagale, R.~T. Bye, R.~Oucheikh, O.~L. Osen, and T.~I. Fossen, ``Path
  planning and collision avoidance for autonomous surface vehicles {II}: a
  comparative study of algorithms,'' \emph{Journal of Marine Science and
  Technology}, vol.~26, no.~4, pp. 1307--1323, 02 2021.

\bibitem{c0071}
C.~Pek and M.~Althoff, ``Efficient computation of invariably safe states for
  motion planning of self-driving vehicles,'' in \emph{2018 IEEE/RSJ
  International Conference on Intelligent Robots and Systems (IROS)}.\hskip 1em
  plus 0.5em minus 0.4em\relax IEEE, Oct. 2018.

\bibitem{c21}
F.~Kr{\"o}ger, \emph{Automated Driving in Its Social, Historical and Cultural
  Contexts}.\hskip 1em plus 0.5em minus 0.4em\relax Berlin, Heidelberg:
  Springer Berlin Heidelberg, 2016, pp. 41--68.

\bibitem{c0060}
D.~Fassbender, B.~C. Heinrich, and H.-J. Wuensche, ``Motion planning for
  autonomous vehicles in highly constrained urban environments,'' in \emph{2016
  {IEEE}/{RSJ} International Conference on Intelligent Robots and Systems
  (IROS)}.\hskip 1em plus 0.5em minus 0.4em\relax IEEE, Oct. 2016.

\bibitem{c0065}
M.~Emam, T.~Rottmann, and M.~Gerdts, ``Efficient real-time obstacle avoidance
  using multi-objective nonlinear model predictive control and semi-smooth
  newton method,'' in \emph{Proceedings of the 10th International Conference on
  Vehicle Technology and Intelligent Transport Systems}.\hskip 1em plus 0.5em
  minus 0.4em\relax SCITEPRESS - Science and Technology Publications, 2024.

\bibitem{c0067}
D.~Lam, C.~Manzie, and M.~Good, ``Model predictive contouring control,'' in
  \emph{49th IEEE Conference on Decision and Control (CDC)}.\hskip 1em plus
  0.5em minus 0.4em\relax IEEE, Dec. 2010.

\bibitem{c0069}
\BIBentryALTinterwordspacing
A.~Romero, S.~Sun, P.~Foehn, and D.~Scaramuzza, ``Model predictive contouring
  control for time-optimal quadrotor flight,'' 2022. [Online]. Available:
  \url{https://arxiv.org/abs/2108.13205}
\BIBentrySTDinterwordspacing

\bibitem{c0014}
M.~Burger and M.~Gerdts, \emph{DAE Aspects in Vehicle Dynamics and Mobile
  Robotics}.\hskip 1em plus 0.5em minus 0.4em\relax Cham: Springer
  International Publishing, 2019, pp. 37--80.

\bibitem{c0072}
C.~Pek and M.~Althoff, ``Computationally efficient fail-safe trajectory
  planning for self-driving vehicles using convex optimization,'' in \emph{2018
  21st International Conference on Intelligent Transportation Systems
  (ITSC)}.\hskip 1em plus 0.5em minus 0.4em\relax IEEE, Nov. 2018, pp.
  1447--1454.

\bibitem{c0063}
M.~Werling, J.~Ziegler, S.~Kammel, and S.~Thrun, ``Optimal trajectory
  generation for dynamic street scenarios in a fren\'{e}t frame,'' in
  \emph{2010 IEEE International Conference on Robotics and Automation}.\hskip
  1em plus 0.5em minus 0.4em\relax IEEE, May 2010.

\bibitem{c0061}
M.~Emam and M.~Gerdts, ``Mpc-based routing and tracking architecture for safe
  autonomous driving in urban traffic,'' \emph{SN Computer Science}, vol.~5,
  no.~4, Mar. 2024.

\bibitem{c0016}
A.~D. Ames, S.~Coogan, M.~Egerstedt, G.~Notomista, K.~Sreenath, and P.~Tabuada,
  ``Control barrier functions: Theory and applications,'' in \emph{2019 18th
  European Control Conference (ECC)}.\hskip 1em plus 0.5em minus 0.4em\relax
  IEEE, June 2019, pp. 3420--3431.

\bibitem{c0076}
Q.~Wang and M.~Gerdts, ``Risk-based path planning for autonomous vehicles,''
  2022.

\bibitem{c0075}
H.~Pan, M.~Luo, J.~Wang, T.~Huang, and W.~Sun, ``A safe motion planning and
  reliable control framework for autonomous vehicles,'' \emph{IEEE Transactions
  on Intelligent Vehicles}, vol.~9, no.~4, pp. 4780--4793, Apr. 2024.

\bibitem{c0059}
D.~Fassbender, A.~Mueller, and H.-J. Wuensche, ``Trajectory planning for
  car-like robots in unknown, unstructured environments,'' in \emph{2014
  {IEEE}/{RSJ} International Conference on Intelligent Robots and
  Systems}.\hskip 1em plus 0.5em minus 0.4em\relax IEEE, Sept. 2014.

\bibitem{c0074}
J.~Pan, P.~Sotiriadis, and F.~Englberger, ``Independent near-field monitoring:
  A novel approach to improve active safety in autonomous vehicles,'' in
  \emph{2024 IEEE Intelligent Vehicles Symposium (IV)}.\hskip 1em plus 0.5em
  minus 0.4em\relax IEEE, June 2024, pp. 730--736.

\bibitem{c0038}
M.~Emam and M.~Gerdts, ``Sensitivity updates for linear-quadratic optimization
  problems in multi-step model predictive control,'' \emph{Journal of Physics:
  Conference Series}, vol. 2514, no.~1, p. 012008, 05 2023.

\bibitem{c0077}
\BIBentryALTinterwordspacing
R.~Schubert, E.~Richter, and G.~Wanielik, ``Comparison and evaluation of
  advanced motion models for vehicle tracking,'' in \emph{2008 11th
  International Conference on Information Fusion}, 2008. [Online]. Available:
  \url{https://api.semanticscholar.org/CorpusID:17357451}
\BIBentrySTDinterwordspacing

\bibitem{c0056}
R.~Graham, ``An efficient algorith for determining the convex hull of a finite
  planar set,'' \emph{Information Processing Letters}, vol.~1, no.~4, pp.
  132--133, June 1972.

\bibitem{c0055}
I.~E. Sutherland and G.~W. Hodgman, ``Reentrant polygon clipping,''
  \emph{Communications of the ACM}, vol.~17, no.~1, pp. 32--42, Jan. 1974.

\bibitem{c0078}
\BIBentryALTinterwordspacing
J.~Zhou, A.~Balachandran, B.~Olofsson, L.~Nielsen, and E.~Frisk, ``Homotopic
  optimization for autonomous vehicle maneuvering,'' in \emph{2024 IEEE
  Intelligent Vehicles Symposium (IV)}.\hskip 1em plus 0.5em minus 0.4em\relax
  IEEE, June 2024, pp. 2561--2568. [Online]. Available:
  \url{http://dx.doi.org/10.1109/IV55156.2024.10588609}
\BIBentrySTDinterwordspacing

\bibitem{c0013}
L.~Gr\"{u}ne and J.~Pannek, \emph{Nonlinear Model Predictive Control: Theory
  and Algorithms}, ser. Communications and Control Engineering.\hskip 1em plus
  0.5em minus 0.4em\relax London, England: Springer, 04 2011.

\bibitem{c0036}
M.~Emam and M.~Gerdts, ``Deterministic operating strategy for multi-objective
  nmpc for safe autonomous driving in urban traffic,'' in \emph{Proceedings of
  the 8th International Conference on Vehicle Technology and Intelligent
  Transport Systems - VEHITS}, 2022, pp. 152--161.

\end{thebibliography}

\end{document}